\documentclass{article}
\pdfoutput=1
\usepackage{arxiv}

\usepackage[utf8]{inputenc} % allow utf-8 input
\usepackage[T1]{fontenc}    % use 8-bit T1 fonts
\usepackage{hyperref}       % hyperlinks
\usepackage{url}            % simple URL typesetting
\usepackage{booktabs}       % professional-quality tables
\usepackage{microtype}      % microtypography
\usepackage{lipsum}		% Can be removed after putting your text content
\usepackage{graphicx}
\usepackage{doi}
\usepackage[section]{placeins}
\usepackage{mathtools}
\usepackage{amsmath}
\usepackage{amssymb}
\usepackage[ruled,vlined]{algorithm2e}
\title{\textbf{Meshless physics-informed deep learning method for three-dimensional solid mechanics}}

\author{{\hspace{1mm}Diab W. Abueidda}\thanks{abueidd2@illinois.edu} \\
	National Center for Supercomputing Applications\\
	Department of Mechanical Science and Engineering\\
	University of Illinois at Urbana-Champaign\\
	\And
    {\hspace{1mm}Qiyue Lu} \\
	National Center for Supercomputing Applications\\
	University of Illinois at Urbana-Champaign\\
	\And
	{\hspace{1mm}Seid Koric}\\
	National Center for Supercomputing Applications\\
	Department of Mechanical Science and Engineering\\
	University of Illinois at Urbana-Champaign\\
}

\date{}

\renewcommand{\shorttitle}{DCM for computational solid mechanics}

\begin{document}

\maketitle

\begin{abstract}
Deep learning and the collocation method are merged and used to solve partial differential equations describing structures' deformation. We have considered different types of materials: linear elasticity, hyperelasticity (neo-Hookean) with large deformation, and von Mises plasticity with isotropic and kinematic hardening. The performance of this deep collocation method (DCM) depends on the architecture of the neural network and the corresponding hyperparameters. The presented DCM is meshfree and avoids any spatial discretization, which is usually needed for the finite element method (FEM). We show that the DCM can capture the response qualitatively and quantitatively, without the need for any data generation using other numerical methods such as the FEM. Data generation usually is the main bottleneck in most data-driven models. The deep learning model is trained to learn the model's parameters yielding accurate approximate solutions. Once the model is properly trained,  solutions can be obtained almost instantly at any point in the domain, given its spatial coordinates. Therefore, the deep collocation method is potentially a promising standalone technique to solve partial differential equations involved in the deformation of materials and structural systems as well as other physical phenomena.
\end{abstract}


\section{Introduction}

Computational solid mechanics aims to predict and/or optimize the behavior of a particular problem using computer methods. Such a problem emerges in natural and engineered systems. Conventional approaches employed to solve partial differential equations (PDEs) governing physical phenomena appearing in computational solid mechanics problems include the isogeometric analysis \cite{hughes2018isogeometric}, the meshfree methods \cite{huerta2018meshfree}, and the finite element method (FEM) \cite{hughes2012finite, ibrahimbegovic2009nonlinear}. Computational methods capable of capturing the physical responses can be computationally expensive and time-consuming. Such problems include and are not limited to multiphysics \cite{zimmerman2006multiphysics, lopeztwo}, modeling of architected materials with complex geometries \cite{abueidda2020compression, babaee20133d, abueidda2018shielding}, nonlinear topology optimization \cite{alberdi2019design, james2015topology, bruns2001topology, jung2004topology}, multiscale analysis \cite{matouvs2017review, mcdowell2009integrated}, etc.

Recently, machine learning (ML) has been proven effective and successful in many fields such as medical diagnoses, image and speech recognition, financial services, autopilot in automotive scenarios, and many other engineering and medical applications \cite{lim2016speech, thorat2019self, DEBRUIJNE201694}. Computational engineering and mechanics are no exception \cite{mielke2019evaluating, ghommem2020fluid, bartovn2020efficient}. Several data-driven approaches have been developed to capture the thermal conductivity of composites \cite{rong2019predicting}, the elastic properties of composites \cite{abueidda2019prediction}, the anisotropic hyperelasticity \cite{VLASSIS2020113299}, the plastic response of different systems \cite{ma2020computational, mozaffar2019deep, settgast2019constitutive, abueidda2020deep}, the thermo-viscoplastic modeling of solidification \cite{abueidda2020deep}, the failure of composites \cite{yang2020prediction}, the fatigue of materials \cite{spear2018data}, the effect of flexoelectricity on nanostructures \cite{hamdia2019novel}, the properties of phononic crystals \cite{sadat2020machine}, etc. Also, several researchers programmed user-defined material subroutines (UMATs) in which neural networks were trained and used to replace the constitutive model integration in typical implicit nonlinear finite element solution procedures \cite{settgast2019constitutive, hashash2004numerical, jung2006neural}. Additionally, data-driven models have shown broad applicability to accelerate design processes and materials discovery \cite{abueidda2019prediction, hamdia2019novel, bessa2019bayesian, chen2020generative, abueidda2020topology, KOLLMANN2020109098}. Recent studies have also shown that deep-learning based surrogate models can establish the material law for composite materials using the computationally or experimentally generated stress-strain data under different loading paths \cite{yang2020learning}.  

While deep learning (DL) provides a platform that is capable of prominently rapid inference, it requires a large training dataset to learn the multifaceted relationships between the inputs and outputs. The dataset size used for training a model is problem-based, i.e., complex problems necessitate large datasets to yield models capable of predicting the response accurately. Thus, such deep-learning based surrogate models usually require a discretization method, such as the finite element method, to generate the data needed to train the model (e.g., \cite{rong2019predicting, abueidda2019prediction, VLASSIS2020113299, ma2020computational, mozaffar2019deep, yang2020prediction, spear2018data, hamdia2019novel, chen2020generative, abueidda2020topology, KOLLMANN2020109098, gu2018novo}). Data-driven approaches are still dominant when it comes to using machine learning algorithms in the field of computational mechanics. Nevertheless, several researchers recently proposed using deep neural networks to directly solve partial differential equations, where such an approach was proposed some time ago \cite{lee1990neural, lagaris1998artificial, lagaris2000neural, malek2006numerical}. However, it did not gain much interest because of the lack of efficient techniques and tools, such as automatic differentiation \cite{baydin2017automatic} and recent advances related to the graphics processing unit (GPU). Raissi et al. \cite{raissi2019physics} developed a physics-informed neural network (PINN) to solve forward and inverse problems. The method has dealt with the strong form, where it is based on a collocation approach. They have used limited data observations that are merged with partial differential equations. The PINN approach \cite{raissi2019physics, guo2020solving, luo2020parameter, haghighat2020deep} has been adopted in several fields, such as cardiovascular flows modeling \cite{kissas2020machine}, poromechanics \cite{kadeethum2020physics}, subsurface transport \cite{he2020physics}, etc.

Solving partial differential equations using deep learning algorithms is very intriguing, and it is gaining an increasing interest by both academia and industry \cite{han2018solving, ling2016reynolds, weinan2017deep, lin2020deep}. Weinan et al. \cite{weinan2018deep} proposed a method called the deep Ritz method, in which they solve variational problems, particularly those arising from partial differential equations. They have shown that the deep Ritz method is naturally nonlinear and adaptive and can be applied to systems with high-dimensional partial differential equations. Additionally, Sirignano \cite{sirignano2018dgm} developed an approach in which the solution is approximated using a deep neural network. Specifically, the proposed meshfree approach trains a deep neural network model to satisfy the differential operator, boundary conditions, and initial conditions.

There are limited attempts to solve partial differential equations related to the field of computational solid mechanics using deep neural networks \cite{samaniego2020energy, nguyen2020deep}. In these attempts, the authors developed a novel approach, called the deep energy method (DEM), where such an approach is well suited for problems possessing energy functionals. In their approach, the potential energy defines the neural network model's loss function, which is minimized using a predefined optimizer, such as the Adam \cite{kingma2014adam} and L-BFGS \cite{liu1989limited} optimizers. The proposed approach has shown promising results when applying it to elasticity, elastodynamics, hyperelasticity, phase field for fracture, and piezoelectricity \cite{samaniego2020energy, nguyen2020deep}.

This study proposes a meshfree approach to solve partial differential equations involving different constitutive models, using deep neural networks (DNNs). In this meshfree approach, the DNN attempts to find the displacement field that satisfies the partial differential equation and the essential and natural boundary conditions. Since the training of a machine learning model is an optimization problem, in which the loss function is minimized, we define the loss function using the strong form. The loss function can be minimized using one of the optimizers, such as the Adam \cite{kingma2014adam} and the quasi-Newton L-BFGS \cite{liu1989limited} optimizers. Here, we refer to this approach as the deep collocation method (DCM). Note that this approach is meshfree and does not require the tangent modulus assembly, which is a primary step in any finite element analysis (FEA) that can be computationally expensive. Additionally, this approach does not necessarily require the definition of the potential energy functional.

The remainder of the paper is laid out as follows. Section \ref{method} talks about the details of the proposed approach, where a generic problem setup is introduced. In Sections \ref{elasticity}, \ref{Hyper}, and \ref{plasticity}, the DCM approach is applied to solve three-dimensional (3D) examples involving elastic, hyperelastic, and elastoplastic constitutive models, respectively. We conclude the paper in Section \ref{conclu} by summarizing the significant results and stating possible future work directions.
\section{Method} \label{method}

When a nonlinear problem involving material and/or geometric nonlinearities is solved using the implicit finite element method, one often ends up defining a residual vector. Using an iterative scheme (e.g., Newton-Raphson), in each iteration, the tangent matrix and residual vector must be obtained to solve the corresponding linear system of equations, using a direct or iterative solver, to find the vector of unknowns. In contrast, explicit nonlinear finite element methods don’t simultaneously solve a linear equations system, but they are often bounded by the conditional stability, often using very small time increments. Moreover, when explicit FEM is used for quasi-static simulations, care must be taken so that the inertia effects are insignificant \cite{koric2009explicit}. This study uses meshfree deep learning to attain the displacement field. Here, the displacement field obtained from the DNN is used to calculate strains, stresses, and other variables needed to satisfy the strong form. The loss function is minimized within a deep learning framework such as PyTorch \cite{NEURIPS2019_9015} and TensorFlow \cite{tensorflow2015-whitepaper}. Below, we include a brief introduction to deep learning and then scrutinize the proposed framework.

\subsection{Introduction to dense neural networks}

\begin{figure}[!htb]
    \centering
    \includegraphics[width=0.65\textwidth]{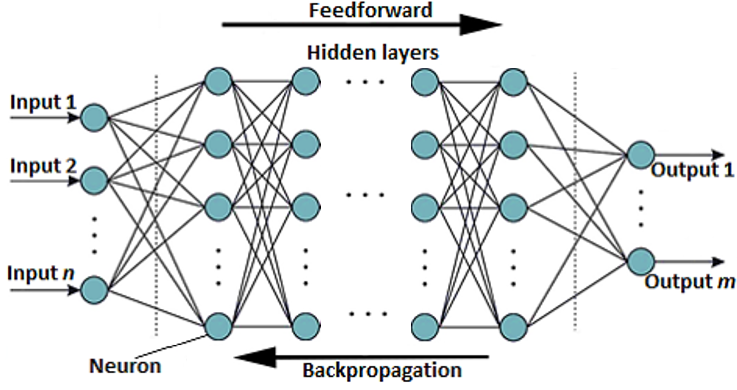}
    \caption{Fully connected (dense) artificial neural network.}
    \label{DANN}
\end{figure}

Deep learning is a special kind of machine learning working with various types of artificial neural networks (ANN) \cite{michalski2013machine}. Loosely inspired by the brain's structure and functionality, artificial neural networks are layers of interconnected individual units, called neurons, where each neuron performs a distinct function given input data. The most straightforward neural network, so-called dense feedforward neural network, comprises linked layers of neurons that map the input data to predictions, as shown in Figure \ref{DANN}. An artificial neural network's deepness is controlled by the number of hidden layers, i.e., the layers in between input and output layers. The numbers of layers and neurons in each layer are specified based on problem complexity. 

\FloatBarrier

The mapping from the input to the output is expressed as $\boldsymbol{Z}: \mathbb{R}^n \rightarrow \mathbb{R}^m$, where $n$ and $m$ denote the number of neurons in the input and output layers, respectively. Upon initialization, the weights $\boldsymbol{W}$ and biases $\boldsymbol{b}$ of the model will be far from ideal. Throughout the training process, the data flow from the model goes forward from the input layer to the output layer of the neural network. The output $\boldsymbol{\hat{o}}^{l}$ of the $l^{th}$ layer is calculated as:
\begin{equation}\label{AEq1}
\begin{aligned}
    \boldsymbol{Z}^{l}&=\boldsymbol{W}^{l} \boldsymbol{\hat{o}}^{l-1}+\boldsymbol{b}^{l}\\
    \boldsymbol{\hat{o}}^{l}&=f^{l}\left(\boldsymbol{Z}^{l}\right){.}\\
\end{aligned}
\end{equation}
After each training pass, $\boldsymbol{W}$ and $\boldsymbol{b}$ are updated. $f^{l}$ represents the activation function, which is an $\mathbb{R} \rightarrow \mathbb{R}$ mapping that acts on the $i^{th}$ neuron in the $l^{th}$ layer. Neural networks employ simple differentiable nonlinear activation functions, which assist the network to learn complex functional relationships between outputs and inputs and render accurate predictions. Some standard activation functions utilized in neural networks are hyperbolic tangent, rectified linear unit (ReLU), sigmoid, etc.

One needs to define a loss function $\mathcal{L}$ and then find the weights $\boldsymbol{W}$ and biases $\boldsymbol{b}$ that lead to a minimized loss value, where such an optimization process is called training in the context of machine learning. The training process requires the use of backpropagation, wherein the loss function is minimized iteratively. One of the most common and most straightforward optimizers used in machine learning is gradient descent, as expressed below:
\begin{equation}\label{AEq2}
\begin{aligned}
    W_{ij}^{c+1}&=W_{ij}^{c}-\beta \frac{\partial \, \mathcal{L}}{\partial W_{ij}^{c}}\\
        b_{i}^{c+1}&=b_{i}^{c}-\beta \frac{\partial \, \mathcal{L}}{\partial b_{i}^{c}}\\
    \end{aligned}
\end{equation}
where $\beta$ denotes the learning rate, which is a crucial parameter in the neural network training process. The gradients of the loss function are obtained using the chain rule. In other words, the gradients of the loss function are calculated with respect to the weights and biases in the last layer, and the weights and biases are updated for each neuron. The same process is then performed for the previous layer and until all of the layers have had their weights and biases updated. Then, a new iteration $c$ with forward propagation starts again. Eventually, after a reasonable number of iterations, the weights $\boldsymbol{W}^{c}$ and biases $\boldsymbol{b}^{c}$ will converge toward a minimized loss value. One also can use more intricate optimizers such as Adam and L-BFGS. Although we use only dense layers and activation functions, the approach is not limited to those; one may incorporate other machine learning algorithms such as convolutional neural network and dropout.

The universal approximation theorem \cite{cybenko1989approximation, hornik1991approximation} can explain why a feedforward neural network can be used as an arbitrary approximator for any continuous function $F\left(\boldsymbol{X}\right)$ defined on a compact subset of $\mathbb{R}^n$. The universal approximation theorem states that the multilayer feedforward neural networks with an arbitrary bounded and nonconstant activation function $f$ and as few as a single hidden layer can be universal approximators with arbitrary accuracy $\epsilon>0$. Providing the activation function is bounded, nonconstant, and continuous, then continuous mappings can be uniformly learned over compact input sets. Mathematically, the universal approximation theorem states that there exist $\boldsymbol{W}$, $\boldsymbol{b}$, $N$, and $\boldsymbol{a}$ such that the approximation function $g\left(\boldsymbol{X}\right)$ satisfies: 
\begin{equation}\label{UAT}
\begin{aligned}
    \left|F\left(\boldsymbol{X}\right)-g\left(\boldsymbol{X}\right)\right|<\epsilon \\
    g\left(\boldsymbol{X}\right)=\sum_{i=1}^{N} a_i f\left(W_{ij} X_j + b_i\right)
\end{aligned}
\end{equation}
where $a_i \in \mathbb{R}$ is fixed. It is noteworthy to highlight that the theorem neither makes conclusions about the network's training, nor the number of neurons needed in the hidden layers to attain the desired accuracy $\epsilon$, nor whether the network's parameters estimation is even feasible.  

\subsection{Deep collocation method}\label{DCM}

Now, we discuss the deep collocation method (DCM), and then, in Section \ref{archit}, we talk about architecture of the neural network model used in this study. In a general sense, the partial differential equation, with solution $\boldsymbol{u}\left(t,\boldsymbol{X}\right)$, can be expressed as: 
\begin{equation}\label{PDEs}
\begin{aligned}
    \left(\partial_t + \mathcal{N}\right)\boldsymbol{u}\left(t,\boldsymbol{X}\right)&=\boldsymbol{0}{,} \quad  \left(t,\boldsymbol{X}\right) \in \left[0,T\right]\times\Omega,\\
    \boldsymbol{u}\left(0,\boldsymbol{X}\right)&=\boldsymbol{u}_o \quad  \boldsymbol{X} \in \Omega,\\
    \boldsymbol{u}\left(t,\boldsymbol{X}\right)&=\overline{\boldsymbol{u}}, \quad \! \left(t,\boldsymbol{X}\right) \in \left[0,T\right]\times\Gamma_{u},\\
    \boldsymbol{t}\left(t,\boldsymbol{X}\right)&=\overline{\boldsymbol{t}}, \quad  \left(t,\boldsymbol{X}\right) \in \left[0,T\right]\times\Gamma_{t},\\
\end{aligned}
\end{equation}
where $\partial_t$ is the partial derivative with respect to time $t$, $T$ is the total time, $\mathcal{N}$ is a spatial differential operator, $\boldsymbol{u}_o$ is the initial condition, $\overline{\boldsymbol{u}}$ is a defined essential boundary condition, and $\overline{\boldsymbol{t}}$ is a defined natural boundary condition. $\Omega$ is the material domain, while $\Gamma_u$ and $\Gamma_t$ are the surfaces with essential and natural boundary conditions, respectively.

In this work, we are  trying to solve partial differential equations by training a neural network with parameters $\boldsymbol {\phi}= \{\boldsymbol{W}, \boldsymbol{b}\}$. Specifically, we train the model such that the approximate solution $\boldsymbol{\hat{u}} \left(t,\boldsymbol{X}; \boldsymbol{\phi} \right)$ obtained from the neural network should be as close as possible to the solution of the differential equation. Hence, the loss function $\mathcal{L}$ is defined using the mean square error ($MSE$) in the light of Equation \ref{PDEs}, i.e.,
\begin{equation}\label{LOSS}
\begin{aligned}
    \mathcal{L}&=MSE_G + \lambda_u\; MSE_u + \lambda_t \; MSE_t + \lambda_i \; MSE_i, \quad \text{where}\\
    MSE_G &= \frac{1}{N_G}\sum_{j=1}^{N_G}\Vert \left(\partial_t + \mathcal{N}\right)\boldsymbol{\hat{u}}\left(t,\boldsymbol{X};\boldsymbol{\phi}\right) \Vert^2{,} \quad \left(t_j,\boldsymbol{X}_j\right) \in \left[0,T\right]\times\Omega\\
    MSE_u &= \frac{1}{N_u}\sum_{j=1}^{N_u}\Vert \boldsymbol{\hat{u}}\left(t,\boldsymbol{X};\boldsymbol{\phi}\right)-\overline{\boldsymbol{u}} \Vert^2{,} \quad  \quad \quad  \; \left(t_j,\boldsymbol{X}_j\right) \in \left[0,T\right]\times\Gamma_u\\
    MSE_t &= \frac{1}{N_t}\sum_{j=1}^{N_t}\Vert \boldsymbol{\hat{t}}\left(t,\boldsymbol{X};\boldsymbol{\phi}\right)-\overline{\boldsymbol{t}} \Vert^2, \quad \quad \quad \; \; \; \left(t_j,\boldsymbol{X}_j\right) \in \left[0,T\right]\times\Gamma_t\\
    MSE_i &= \frac{1}{N_i}\sum_{j=1}^{N_i}\Vert \boldsymbol{\hat{u}}\left(0,\boldsymbol{X};\boldsymbol{\phi}\right)-\boldsymbol{u}_o \Vert^2{,} \quad \quad \quad \boldsymbol{X}_j \in \Omega\\
\end{aligned}
\end{equation}
where $N_G$, $N_u$, $N_t$, and $N_i \in \Omega$ are the numbers of the sampled points corresponding to the different terms of the loss function. $\lambda_u>0$, $\lambda_t>0$, and $\lambda_i>0$ are hyperparameters (penalty coefficients) weighting the contribution of the different boundary conditions. In our experience, these weighting hyperparameters are essential to guarantee that the different terms are contributing to a similar degree to the value of the loss function.

\begin{figure}[!htb]
    \centering
    \includegraphics[width=0.8\textwidth]{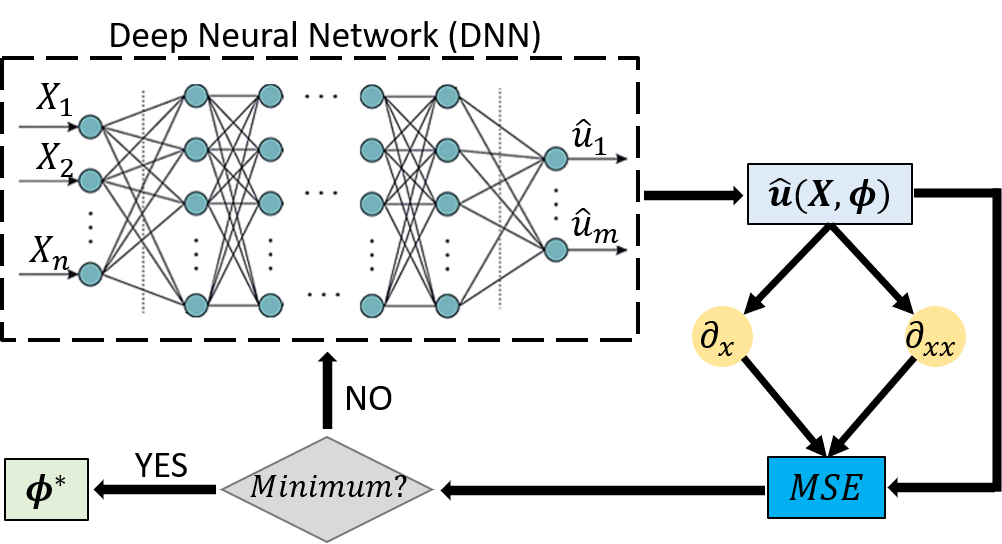}
    \caption{Flow chart of the deep collocation method (DCM).}
    \label{DCM_FC}
\end{figure}

In the examples we have considered in this study, we assume that the inertia term is zero, i.e., the term $\partial_t \boldsymbol{u} = \boldsymbol{0}$, and the solution is $\boldsymbol{u} \left( \boldsymbol{X} \right)$. Hence, there is no need to define initial conditions, i.e., the $MSE_i$ term drops from the definition of the loss function. Figure \ref{DCM_FC} depicts the deep collocation method used here. The deep neural network (DNN) minimizes the loss function to obtain the optimized network parameters $\boldsymbol{\phi}^{*} =\{\boldsymbol{W}^*, \boldsymbol{b}^*\}$. The DNN model maps the coordinates $\boldsymbol{X}$ of the sampled points to the displacement field $\boldsymbol{\hat{u}}\left(\boldsymbol{X}, \boldsymbol{\phi}\right)$ using the feedforward propagation. The predicted displacement field $\boldsymbol{\hat{u}}$ is used to calculate the dependent variables, involved in a specific problem, and the loss function. The computation of the dependent variables and loss function typically requires finding the first and second derivative of $\boldsymbol{\hat{u}}$, which are found using the automatic differentiation offered by deep learning frameworks. The optimization (minimization) problem is written as:
\begin{equation}\label{BC_NN}
\begin{aligned}
    \boldsymbol{\phi}^{*}&=\underset{\boldsymbol{\phi}}{\mathrm{arg\,min}} \; \mathcal{L}\left(\boldsymbol{\phi}\right){.}\\
\end{aligned}
\end{equation}

The optimization problem is solved using an optimizer such as L-BFGS, where the gradients are calculated using the backpropagation process. Note that the DCM is a meshfree approach, and there is no need to assemble the tangent modulus, and also there is no linear system of equations has to be solved. Such a method does not necessitate the definition of potential energies, making it intriguing for a broader spectrum of problems, such as plasticity. It is worth mentioning that the optimization problem involved in any machine learning problem, including the DCM, is generally a nonconvex one. Hence, one needs to pay attention to the possibility of getting trapped in local minima and saddle points. This study is far from being the last word on the topic.

\subsection{Design of experiments}\label{DOE}

The design of experiments (DOE) has become prominent in the era of the swiftly growing fields of data science, machine learning, and statistical modeling. The proposed approach is meshfree, in which many points are sampled from three sets, namely $\Omega$, $\Gamma_u$, and $\Gamma_t$. This set of randomly distributed collocation points are used to minimize the loss function while accounting for a set of constraints (boundary conditions). These collocation points are used to train the machine learning model, i.e., to solve the partial differential equations in the physical domain. There are many sampling methods that can be adopted. Here, we use the Monte Carlo method, i.e., create the points through repeated random sampling \cite{shapiro2003monte}.

\subsection{DNN model}\label{archit}

\begin{figure}[!htb]
    \centering
    \includegraphics[width=0.5\textwidth]{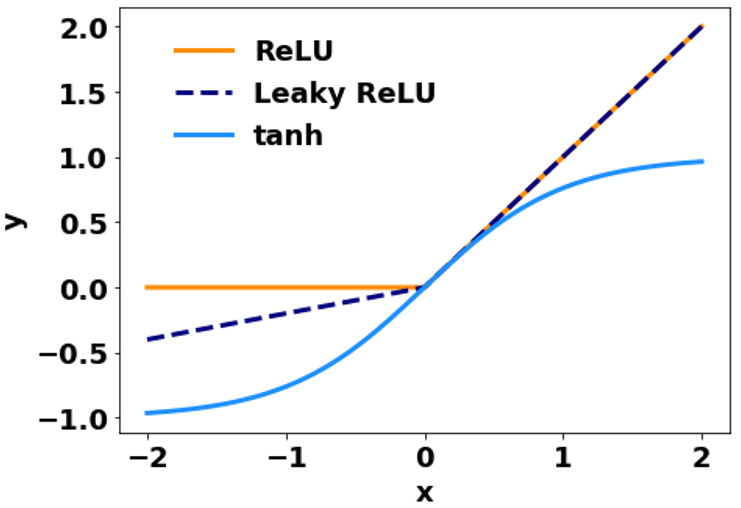}
    \caption{Illustration of some activation functions.}
    \label{Activation}
\end{figure}

In this paper, a $6$-layer network is used, where the numbers of neurons in the different layers are $3-60-60-60-60-3$. Three neurons are used in the input layer for the coordinates $\boldsymbol{X}$, and three neurons are used in the output layer for the displacement $\boldsymbol{\hat{u}}$. In the four hidden layers, the number of neurons employed is $60$. After each dense layer, we use an activation function, as demonstrated in Equation \ref{AEq1}. A network based on dense layers without nonlinear activation functions is reduced to a linear one, making it challenging to capture nonlinear relationships between input and output. One popular activation functions in many machine learning practices are ReLU and leaky ReLU. However, this approach requires finding the second-order derivative. While ReLU is a globally nonlinear activation function, it is linear in a neighborhood of almost every input, rendering the network linear in a neighborhood of the inputs (see Figure \ref{Activation}). Hence, it is not a good choice for our case. Here, we use the hyperbolic tangent (tanh) activation function within the hidden layers, where, for the output layer, we use a linear activation function. The architecture of the network is obtained by trying several architectures and fine-tuning of hyperparameters.

As mentioned earlier, the loss function, defined in Equation \ref{LOSS}, is minimized. PyTorch is used in this study to solve the deep learning problem (minimize the loss function). Here, we use two optimizers; we start with the Adam optimizer, and then the limited-memory BFGS (L-BFGS) optimizer with the Strong Wolfe line search \cite{liu1989limited, lewis2013nonsmooth} is used. High-throughput computations are done on iForge, which is an HPC cluster hosted at the National Center for Supercomputing Applications (NCSA). A representative compute node on iForge has the following hardware specifications: two NVIDIA Tesla V100 GPUs, two $20$-core Intel Skylake CPUs, and $192$ GB main memory. Since it is challenging to find an analytical solution for most cases, we obtain a reference solution using the conventional finite element analysis to measure the accuracy of the solution obtained from the DCM. We use the normalized ``$L_2$-error" as a metric reflecting on the accuracy of the DCM solution:
\begin{equation}\label{L2error}
\begin{aligned}
   L_2\text{-error}&=\frac{||\boldsymbol{u}_{REF}-\hat{\boldsymbol{u}}||^{L_{2}}}{||\boldsymbol{u}_{REF}||^{L_{2}}}{.}\\
\end{aligned}
\end{equation}
\section{Elasticity}\label{elasticity}

Here, we consider a homogeneous, isotropic, elastic body going small deformation. The equilibrium equation, in the absence of body and inertial forces, is expressed as:
\begin{equation}\label{eqlbrm}
\begin{split}
    \boldsymbol{\nabla}\cdot \boldsymbol{\sigma}&=\boldsymbol{0}{,} \quad  \boldsymbol{x}\in\Omega,\\
    \boldsymbol{\hat{u}}&=\overline{\boldsymbol{u}}, \quad \! \boldsymbol{x}\in\Gamma_{u},\\
    \boldsymbol{\sigma}\cdot\boldsymbol{n}&=\overline{\boldsymbol{t}}, \quad  \boldsymbol{x}\in\Gamma_{t},\\
\end{split}
\end{equation}
where $\boldsymbol{\sigma}$ is the Cauchy stress tensor, $\boldsymbol{n}$ is the normal unit vector, $\boldsymbol{\nabla}\cdot$ denotes the divergence operator, and $\boldsymbol{\nabla}$ is the gradient operator. Since small deformation is assumed, the strain is given by:
\begin{equation}\label{eps}
\begin{aligned}
    \boldsymbol{\varepsilon}=\frac{1}{2}(\boldsymbol{\nabla} \boldsymbol{\hat{u}}+\boldsymbol{\nabla} \boldsymbol{\hat{u}}^T)\\
    \end{aligned}
\end{equation}
where $u_{i}$ are displacement components, and $\boldsymbol{\varepsilon}$ denotes the infinitesimal strain tensor. The relationship between the stress and strain is expressed as:
\begin{equation}\label{hooks}
\begin{aligned}
    \boldsymbol{\sigma}&=\lambda \; \text{trace}\left(\boldsymbol{\varepsilon}\right) \;\boldsymbol{I} + 2\; \mu \; \boldsymbol{\varepsilon} \\
    \end{aligned}
\end{equation}
where $\lambda$ and $\mu$ denote the Lame constants, and $\boldsymbol{I}$ represents the second-order identity tensor. The steps used to solve a problem with a linear elastic constitutive model are summarized in Algorithm \ref{algo_elast}.

\begin{algorithm}[!htb]
\SetAlgoLined
    \textbf{Input}: Physical domain, BCs, and DNN\\
    \quad \quad Material parameters ($\lambda$ and $\mu$)\\ 
    \quad \quad Sample points $\boldsymbol{X}_{int}$ from $\Omega$\\
    \quad \quad Sample points $\boldsymbol{X}_{u}$ from $\Gamma_u$\\
    \quad \quad Sample points $\boldsymbol{X}_{t}$ from $\Gamma_t$\\
    \quad \quad Neural network architecture\\ 
    \quad \quad Neural network hyperparameters\\
    \quad \quad Optimizer (Adam followed by L-BFGS)\\
    \textbf{Initialization}: Initial weights and biases of the DNN\\
    \textbf{Output}: Optimized weights and biases of the DNN\\
    
 \While{Not minimized}{
    Obtain $\boldsymbol{\hat{u}}$ from the DNN\\
    Compute $\boldsymbol{\nabla} \boldsymbol{\hat{u}}$ using automatic differentiation\\
    Compute $\boldsymbol{\varepsilon}$\\
    Compute $\boldsymbol{\sigma}$\\
    \uIf{$\boldsymbol{X}_{int}$}{
    Compute $\boldsymbol{\nabla}\cdot \boldsymbol{\sigma}$ using automatic differentiation\\
    Calculate $MSE_G$\\
    }
    \uElseIf{$\boldsymbol{X}_{t}$}{
    Compute $\boldsymbol{t}= \boldsymbol{\sigma}\cdot\boldsymbol{n}$\\
    Calculate $MSE_t$\\
    }
    \uElse{
    Calculate $MSE_u$\\
    }
    Calculate loss function\\
    Update the weights and biases\\
 }
\caption{Linear elasticity}
\label{algo_elast}
\end{algorithm}

\begin{figure}[!htb]
    \centering
    \includegraphics[width=0.8\textwidth]{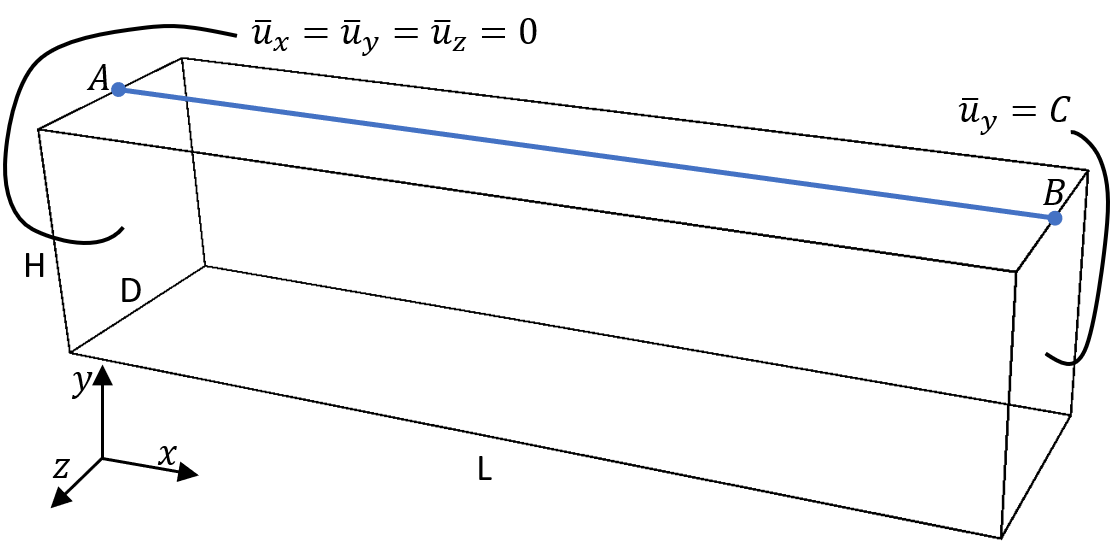}
    \caption{3D beam bending test.}
    \label{Can_BM}
\end{figure}

\begin{figure}[!htb]
    \centering
    \includegraphics[width=0.5\textwidth]{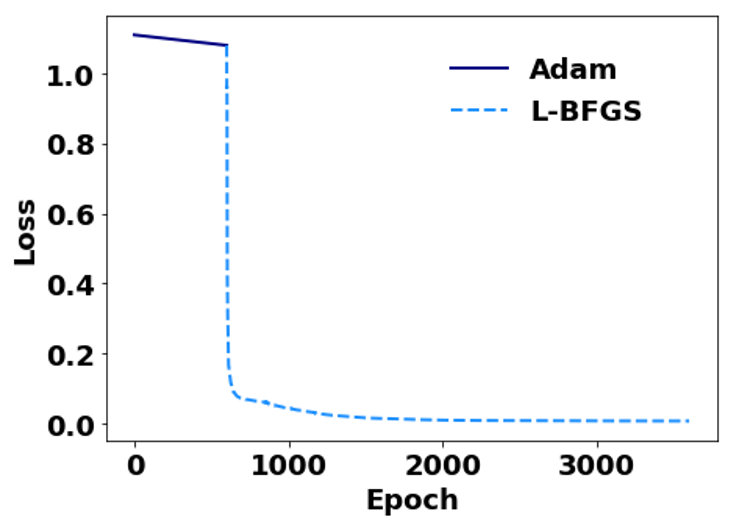}
    \caption{Elasticity example: Loss function convergence.}
    \label{loss_history_elast}
\end{figure}

\begin{figure}[!htb]
    \centering
    \includegraphics[width=1\textwidth]{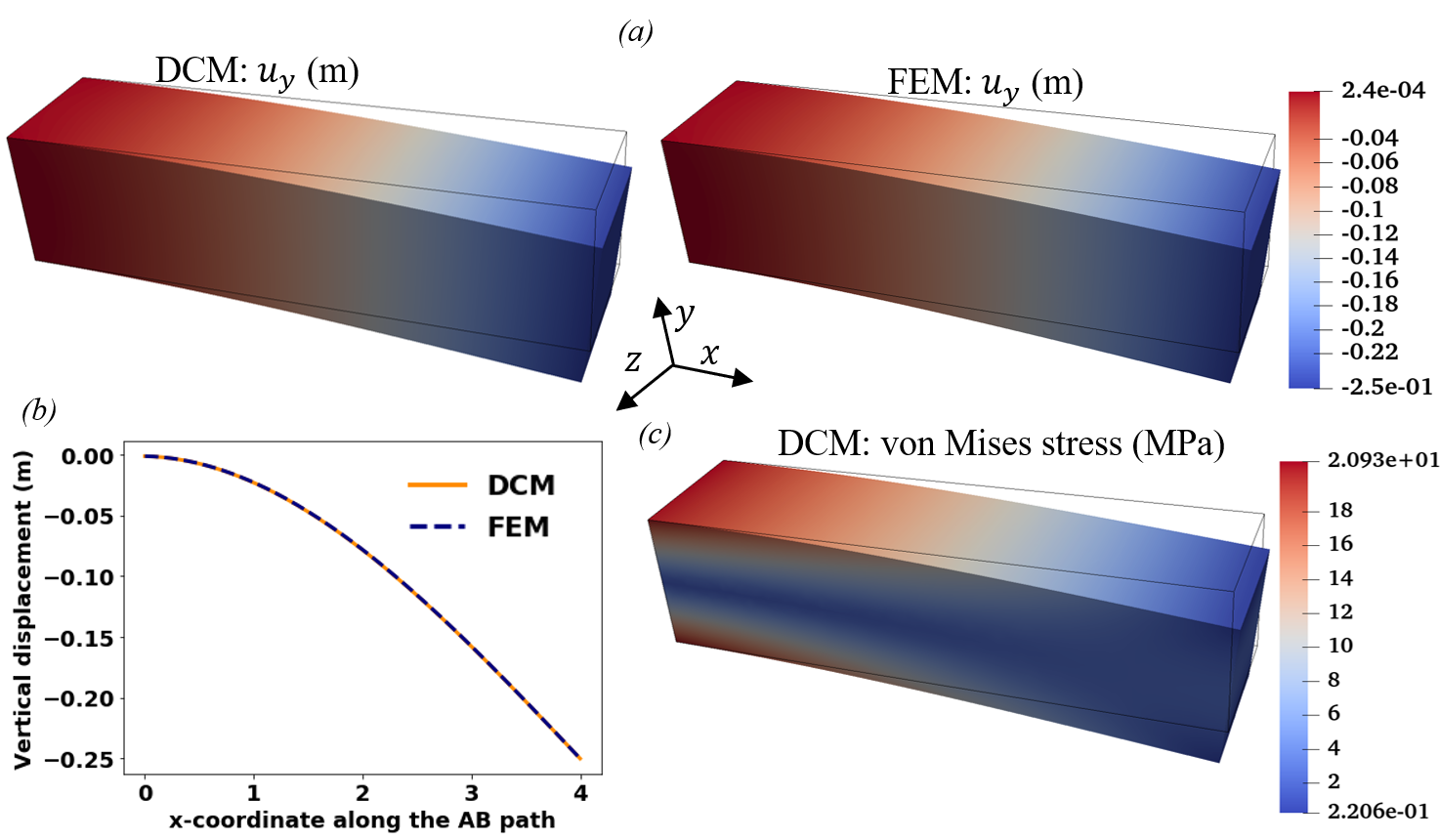}
    \caption{Elasticity example: (a) Comparison between the vertical displacement $\hat{u}_y$ contours obtained from the DCM and FEM, (b) Comparison between the FEM and DCM vertical displacement $\hat{u}_y$ along the $AB$ path, and (c) von Mises stress contours.}
    \label{ElasticityCB}
\end{figure}

Now, let us consider a 3D bending beam, as shown in Figure \ref{Can_BM}. Let $L = 4\;\text{m}$ and $D=H=1\;\text{m}$. Using a displacement-controlled approach, the displacement in the $y$-direction is $\overline{u}_y=C=0.25\;\text{m}$ at the face with a normal unit vector $\boldsymbol{n}=\left[1,0,0\right]^T$. On the other hand, the displacements $\overline{u}_x$, $\overline{u}_y$, and $\overline{u}_z$ are set zero at the face with a normal unit vector $\boldsymbol{n} = \left[-1,0,0\right]^T$. Since this approach is based on the strong form, the degrees of freedom on the boundaries with both zero and nonzero tractions have to be explicitly satisfied. The $MSE_t$ term appearing in Equation \ref{LOSS} takes care of this. The numbers of points sampled from the cantilever beam domain are as follows: $N_G=7500$, $N_u=4000$, and $N_t=4000$. Figure \ref{loss_history_elast} shows the convergence of the loss function. As mentioned earlier, we have used two optimizers. We start with the Adam optimizer, and then it is followed by the L-BFGS optimizer. We found that combining both optimizers stabilizes the optimization procedure. Figure \ref{ElasticityCB} depicts the vertical displacement $\hat{u}_y$ contours obtained from the DCM and compares the contours with those obtained from the FEM. Also, the von Mises stress contours obtained using the DCM are presented. After the model is trained, and the optimized weights and biases are attained, 1000 test points from the DCM, different from the ones used in the training of the DNN, and their corresponding ones from the FEM are utilized to calculate the $L_2\text{-error}$; the $L_2\text{-error}=0.11$. 

Since the constitutive model here is path-independent, the solution can be obtained using one pseudo-time step. Also, although we are sampling the points at the very beginning of the solution procedure and fix those points during the optimization process, one is not really required to do so. In other words, this case is path-independent; therefore, one can sample new points at each optimization iteration, which might help increase the model's accuracy. This is not the case for elastoplastic problems, as discussed later. We sample the points at the beginning of the optimization procedure for simplicity, and we fix them throughout the optimization iteration.
\section{Hyperelasticity}\label{Hyper}
Let us consider a body made of a homogeneous and isotropic hyperelastic material under finite deformation. The mapping $\boldsymbol{\zeta}$ of material points from the initial configuration to the current configuration is given by:
\begin{equation}\label{mapping}
\begin{aligned}
    \boldsymbol{x}&=\boldsymbol{\zeta}\left(\boldsymbol{X},t\right)=\boldsymbol{X}+\boldsymbol{\hat{u}}.\\
\end{aligned}
\end{equation}
In the absence of body and inertial forces, the strong form is written as:
\begin{equation}\label{eqlbrm}
\begin{split}
    \boldsymbol{\nabla}_{\boldsymbol{X}}\cdot \boldsymbol{P}&=\boldsymbol{0}{,} \quad  \boldsymbol{X}\in\Omega,\\
    \boldsymbol{\hat{u}}&=\overline{\boldsymbol{u}}, \quad \! \boldsymbol{X}\in\Gamma_{u},\\
    \boldsymbol{P}\cdot\boldsymbol{N}&=\overline{\boldsymbol{t}}, \quad  \boldsymbol{X}\in\Gamma_{t},\\
\end{split}
\end{equation}
where $\boldsymbol{P}$ is the first Piola-Kirchhoff stress, and $\boldsymbol{N}$ represents the outward normal unit vector in the initial configuration. The constitutive law for such a material is expressed as:
\begin{equation}\label{constit_hyper}
\begin{aligned}
    \boldsymbol{P}&=\frac{\partial \psi\left(\boldsymbol{F}\right)}{\partial \boldsymbol{F}}\\
    \boldsymbol{F}&=\boldsymbol{\nabla}_{\boldsymbol{X}} \boldsymbol{\zeta}\left(\boldsymbol{X}\right)\\
\end{aligned}
\end{equation}
where $\boldsymbol{F}$ denotes the deformation gradient. For the neo-Hookean hyperelastic material, the Helmholtz free energy $\psi\left(\boldsymbol{F}\right)$ is given by:
\begin{equation}\label{neohookenergy}
\begin{aligned}
    \psi\left(\boldsymbol{F}\right)&=\frac{1}{2}\;\lambda\; \left(\text{ln}\left(J\right)\right)^{2}-\mu \; \text{ln}\left(J\right)+\frac{1}{2} \; \mu \; \left(I_{1} -3\right),\\
\end{aligned}
\end{equation}
where the first invariant is defined as $I_{1} = \text{trace} \left(\boldsymbol{C} \right)$, the right Cauchy-Green tensor $\boldsymbol{C}$ is defined as $\boldsymbol{C} =\boldsymbol{F}^{T} \boldsymbol{F}$, and the second invariant is defined as $J=\text{det}\left(\boldsymbol{F}\right)$. Thus, $\boldsymbol{P}$ is given by:
\begin{equation}\label{constit_neo}
\begin{aligned}
    \boldsymbol{P}&=\frac{\partial \psi\left(\boldsymbol{F}\right)}{\partial \boldsymbol{F}}=\mu \; \boldsymbol{F}+\left(\lambda\;\text{ln}\left(J\right)-\mu \right)\;\boldsymbol{F}^{-T},\\
    \boldsymbol{P}&=J\;\boldsymbol{\sigma}\;\boldsymbol{F}^{-T}.\\
\end{aligned}
\end{equation}
\begin{algorithm}[!htb]
\SetAlgoLined
    \textbf{Input}: Physical domain, BCs, and DNN\\
    \quad \quad Material parameters ($\lambda$ and $\mu$)\\ 
    \quad \quad Sample points $\boldsymbol{X}_{int}$ from $\Omega$\\
    \quad \quad Sample points $\boldsymbol{X}_{u}$ from $\Gamma_u$\\
    \quad \quad Sample points $\boldsymbol{X}_{t}$ from $\Gamma_t$\\
    \quad \quad Neural network architecture\\ 
    \quad \quad Neural network hyperparameters\\
    \quad \quad Optimizer (Adam followed by L-BFGS)\\
    \textbf{Initialization}: Initial weights and biases of the DNN\\
    \textbf{Output}: Optimized weights and biases of the DNN\\
    
 \While{Not minimized}{
    Obtain $\boldsymbol{\hat{u}}$ from the DNN\\
    Compute $\boldsymbol{\nabla}_{\boldsymbol{X}} \boldsymbol{\hat{u}}$ using automatic differentiation\\
    Compute $\boldsymbol{F}=\boldsymbol{I}+\boldsymbol{\nabla}_{\boldsymbol{X}} \boldsymbol{\hat{u}}$\\
    Compute $J=\text{det}\left(\boldsymbol{F}\right)$, $\boldsymbol{C}$, $I_1$, and $\boldsymbol{P}$\\
    \uIf{$\boldsymbol{X}_{int}$}{
    Compute $\boldsymbol{\nabla}_{\boldsymbol{X}}\cdot \boldsymbol{P}$ using automatic differentiation\\
    Calculate $MSE_G$\\
    }
    \uElseIf{$\boldsymbol{X}_{t}$}{
    Compute $\boldsymbol{t}= \boldsymbol{P}\cdot\boldsymbol{N}$\\
    Calculate $MSE_t$\\
    }
    \uElse{
    Calculate $MSE_u$\\
    }
    Calculate loss function\\
    Update the weights and biases\\
 }
\caption{neo-Hookean hyperelasticity}
\label{algo_hyperelast}
\end{algorithm} 

The steps used to solve a problem with a neo-Hookean hyperelastic constitutive model are described in Algorithm \ref{algo_hyperelast}. Similar procedure, with slight changes, can be used for other hyperelastic constitutive models, such as the Mooney-Revlin and Arruda-Boyce hyperelastic models. 

\begin{figure}[!htb]
    \centering
    \includegraphics[width=1\textwidth]{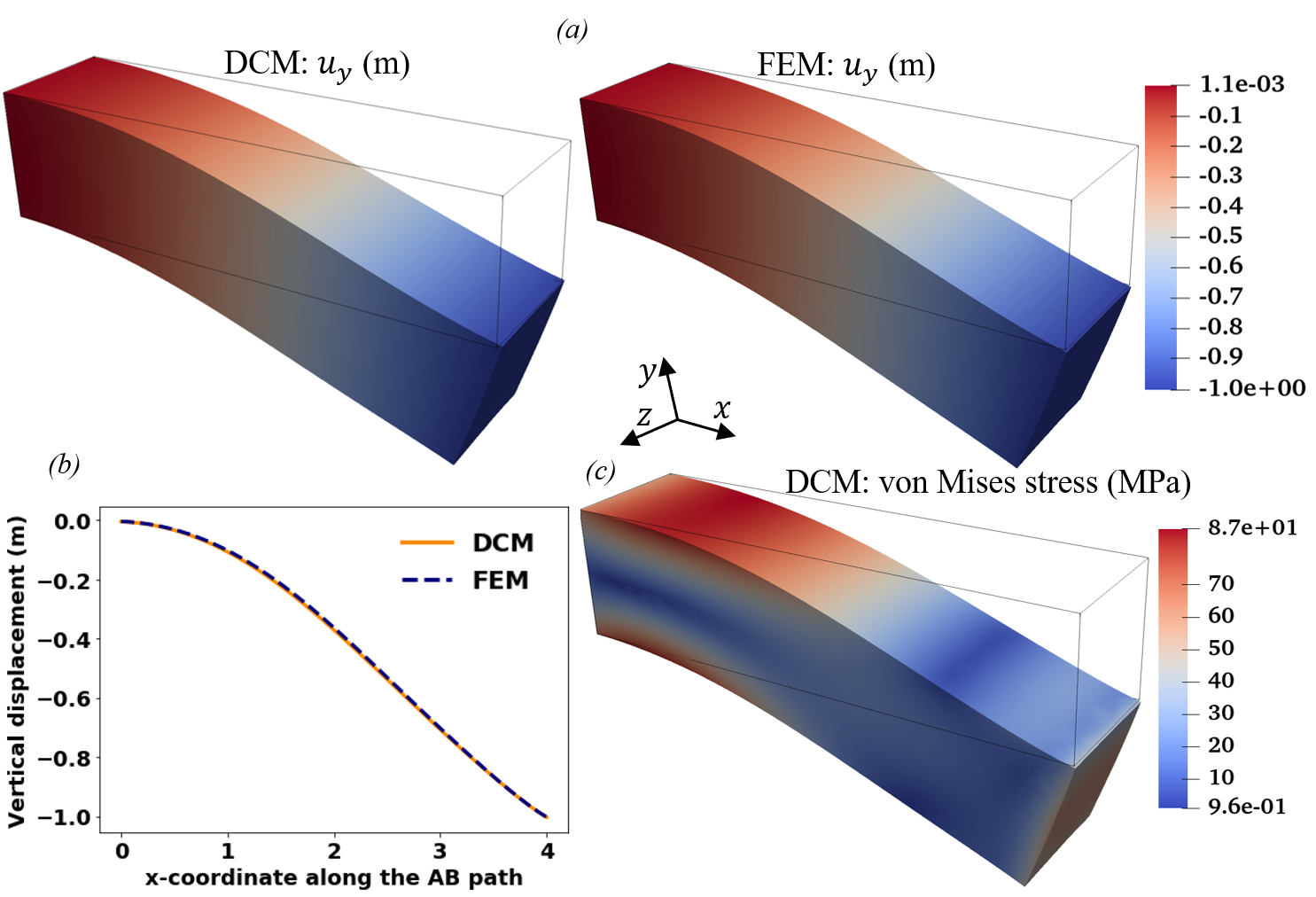}
    \caption{Hyperelasticity example: (a) Comparison between the vertical displacement $\hat{u}_y$ contours obtained from the DCM and FEM, (b) Comparison between the DCM and FEM vertical displacement $\hat{u}_y$ along the $AB$ path, and (c) von Mises stress contours.}
    \label{HyperelasticityCB}
\end{figure}

Going back to the cantilever beam example (see Figure \ref{Can_BM}), but now assume that the cantilever beam is made of a neo-Hookean hyperelastic material. The same boundary conditions are applied, except the vertical displacement $\overline{u}_y$ at the face with the normal unit vector $\boldsymbol{N}=[1,0,0]^T$ is increased to $C=1.0\; \text{m}$ to ensure that large deformation is induced. The numbers of points sampled from the cantilever beam domain are as follows: $N_G=7500$, $N_u=4000$, and $N_t=4000$. Figure \ref{HyperelasticityCB} illustrates the vertical displacement $\hat{u}_y$ contours found using the DCM and compares the contours with those obtained from the FEM. Additionally, the von Mises stress contours obtained using the DCM are shown. Similar to the linear elasticity case, after optimizing the weight and biases of the DNN, 1000 test points from the DCM, different from the ones used in the training of the DNN, and their FEM corresponding ones are used to find the $L_2\text{-error}$; the $L_2\text{-error}=0.27$. When the FEM is used to solve a hyperelastic problem involving large deformation, one usually needs to implement pseudo-time discretization to avoid divergence, although the problem is history-independent. Using the DCM, we managed to solve the problem using one pseudo-time step. Of course, one can find the solution at the intermediate pseudo-time steps if it is of interest. Also, we have not studied the effect of obtaining the solution using multiple pseudo-time steps. It could help increase the accuracy of the obtained solution, but it is not required, as we have shown. Furthermore, similar to the elasticity case, we are sampling the points at the very beginning of the solution procedure, and then those points stay unchanged during the optimization process. Nevertheless, one is not obligated to stick with the same points when the material is hyperelastic, and these points can be resampled at each optimization iteration. We anticipate that resampling the points at each optimization iteration would increase the model's accuracy.

\section{Plasticity}\label{plasticity}

In the second case presented in this study, we consider a body made of a material obeying a $J_{2}$ flow theory with linear isotropic and kinematic hardening, where both forms of the hardening law evolve linearly. A cantilever beam problem is solved here as an example of applying the DCM to elastoplastic systems. The strong form, in the absence of inertial and body forces, is given by:
\begin{equation}\label{eqlbrm_case3}
\begin{split}
    \boldsymbol{\nabla}\cdot \boldsymbol{\sigma}&=\boldsymbol{0}{,} \quad  \boldsymbol{x}\in\Omega,\\
    \boldsymbol{\hat{u}}&=\overline{\boldsymbol{u}}, \quad \! \boldsymbol{x}\in\Gamma_{u},\\
    \boldsymbol{\sigma}\cdot\boldsymbol{n}&=\overline{\boldsymbol{t}}, \quad  \boldsymbol{x}\in\Gamma_{t}{.}\\
\end{split}
\end{equation}
Under the assumption of small deformation, the strain tensor $\boldsymbol{\varepsilon}$ is additively decomposed into the elastic strain tensor $\boldsymbol{\varepsilon}^{el}$ and plastic strain tensor $\boldsymbol{\varepsilon}^{pl}$:
\begin{equation}\label{eps_dec}
\begin{aligned}
    \boldsymbol{\varepsilon}&=\frac{1}{2}\left(\boldsymbol{\nabla}\boldsymbol{u}+\left(\boldsymbol{\nabla}\boldsymbol{u}\right)^{T}\right)\\
    \boldsymbol{\varepsilon}&=\boldsymbol{\varepsilon}^{el}+\boldsymbol{\varepsilon}^{pl}{.}\\
    \end{aligned}
\end{equation}
The von Mises yield condition $y$ is used:
\begin{equation}\label{yield_con}
\begin{aligned}
    y\left(\boldsymbol{\sigma}, \boldsymbol{q}, \alpha\right)&= ||\boldsymbol{\eta}||^f-\sqrt{\frac{2}{3}}\left(\sigma_y+K \alpha\right)\\
    \boldsymbol{\eta}&=\boldsymbol{s}-\boldsymbol{q}\\
\end{aligned}
\end{equation}
where $||\cdot||^f$ is the Frobenius norm, $\boldsymbol{\eta}$ is the relative stress tensor, $\boldsymbol{s}$ is the deviatoric stress tensor, $\boldsymbol{q}$ denotes the back stress, and $\alpha$ is the internal plastic variable known as the equivalent plastic strain. $\sigma_y$ and $K$ are material hardening constants. The Karush-Kuhn-Tucker (KKT) conditions and consistency condition are required to complete the definition of the constitutive model:
\begin{equation}\label{KKT}
\begin{aligned}
    y\left(\boldsymbol{\sigma}, \boldsymbol{q}, \alpha\right) &\leq 0\\
    \gamma y\left(\boldsymbol{\sigma}, \boldsymbol{q}, \alpha\right) &= 0\\
    \gamma &\geq 0\\
    \gamma \dot{y}\left(\boldsymbol{\sigma}, \boldsymbol{q}, \alpha\right)&=0\\
\end{aligned}
\end{equation}
where $\gamma$ is the consistency parameter. The evolution of $\alpha$ and $\boldsymbol{q}$ is defined as:
\begin{equation}\label{evolution}
\begin{aligned}
    \dot{\alpha}&= \sqrt{\frac{2}{3}} \gamma\\
    \boldsymbol{n}&= \frac{\boldsymbol{\eta}}{||\boldsymbol{\eta}||^f}\\
    \dot{\boldsymbol{q}}&=\frac{2}{3} \gamma H \boldsymbol{n}\\
\end{aligned}
\end{equation}
where $H$ is the material kinematic hardening constant, and $\boldsymbol{n}$ is the tensor normal to the yield surface.

The radial return mapping algorithm, originally presented by Wilkins \cite{wilkins1964methods}, is used here. Given the state at an integration point: $\alpha_t$ and $\boldsymbol{\varepsilon}_t^p$ at the previous time step $t$, and $\boldsymbol{\varepsilon}_{t+1}$ at the current time step $t+1$, one can calculate $\alpha_{t+1}$ and $\boldsymbol{\varepsilon}_{t+1}^p$. Firstly, the trial state is computed:
\begin{equation}\label{trial_stress}
\begin{aligned}
    \boldsymbol{e}_{t+1}&=\boldsymbol{\varepsilon}_{t+1}-\frac{1}{3}\,\text{trace}\left(\boldsymbol{\varepsilon}_{t+1}\right)\boldsymbol{I}\\
    \boldsymbol{s}_{t+1}^{trial}&=2\,\mu \left(\boldsymbol{e}_{t+1}-\boldsymbol{e}_{t}^{p}\right)\\
    \boldsymbol{\eta}_{t+1}^{trial}&=\boldsymbol{s}_{t+1}^{trial}-\boldsymbol{q}_{t}\\
    \end{aligned}
\end{equation}
where $\boldsymbol{I}$ is the second-order identity tensor, $\mu$ represents the shear modulus, and $\boldsymbol{e}$ is the deviatoric strain tensor.

Then, the yield condition $y_{t+1}^{trial}$ is checked:
\begin{equation}\label{yield_con_trial}
\begin{aligned}
    y_{t+1}^{trial}&= ||\boldsymbol{\eta}^{trial}_{t+1}||^f-\sqrt{\frac{2}{3}}\left(\sigma_y+K \alpha_t\right)\\    
\end{aligned}
\end{equation}
If $y_{t+1}^{trial}\leq 0$, $\left(\cdot\right)_{t+1}=\left(\cdot\right)^{trial}_{t+1}$. Otherwise, we proceed with the calculations of $\boldsymbol{n}_{t+1}$, $\Delta\boldsymbol{\gamma}_{t+1}$, and $\alpha_{t+1}$:
\begin{equation}\label{plastic_if}
\begin{aligned}
    \boldsymbol{n}_{t+1}&= \frac{\boldsymbol{\eta}^{trial}_{t+1}}{||\boldsymbol{\eta}^{trial}_{t+1}||^f}\\
    \Delta \gamma_{t+1}&=\frac{y_{t+1}^{trial}}{2\left(\mu+\frac{H}{3}+\frac{K}{3}\right)}\\
    \alpha_{t+1}&=\alpha_t + \sqrt{\frac{2}{3}} \Delta \gamma_{t+1}{.}\\
\end{aligned}
\end{equation}
Next, the back stress, plastic strain, and stress are updated:
\begin{equation}\label{update}
\begin{aligned}
    \boldsymbol{q}_{t+1}&=\boldsymbol{q}_{t}+\frac{2}{3}\Delta \gamma_{t+1} \, H\, \boldsymbol{n}_{t+1}\\
    \boldsymbol{e}^{p}_{t+1}&=\boldsymbol{e}^{p}_{t}+\Delta \gamma_{t+1} \boldsymbol{n}_{t+1}\\
    \boldsymbol{\sigma}_{t+1}&=\kappa\, \text{trace}\left(\boldsymbol{\varepsilon}_{t+1}\right)\boldsymbol{I}+\boldsymbol{s}_{t+1}^{trial}-2\,\mu\,\Delta \gamma_{t+1} \boldsymbol{n}_{t+1}\\
\end{aligned}
\end{equation}
where $\kappa$ is the bulk modulus. The procedure used to solve the elastoplastic problem is described in Algorithm \ref{algo_plast}. For elastoplastic problems, the solution is obtained using $M$ pseudo-time steps. Hence, the optimization problem is solved $M$ times, where the optimized weights and biases attained at a step $t$ are used as the initial weights and biases for the next step $t+1$, which can be considered a form of transfer learning. This transfer learning is equivalent to updating displacements after each converged step in a typical nonlinear implicit finite element analysis solution procedure.

\begin{algorithm}[!htb]
\SetAlgoLined
    \textbf{Input}: Physical domain, BCs, and DNN\\
    \quad \quad Material parameters ($\kappa$, $\mu$, $\sigma_y$, $K$, and $H$)\\ 
    \quad \quad Sample points $\boldsymbol{X}_{int}$ from $\Omega$\\
    \quad \quad Sample points $\boldsymbol{X}_{u}$ from $\Gamma_u$\\
    \quad \quad Sample points $\boldsymbol{X}_{t}$ from $\Gamma_t$\\
    \quad \quad Neural network architecture\\ 
    \quad \quad Neural network hyperparameters\\
    \quad \quad Optimizer (Adam followed by L-BFGS)\\
    \quad \quad Initialize $\boldsymbol{e}_{o}^{p}$, $\boldsymbol{q}_{o}$, and $\alpha_o$\\
    \textbf{Initialization}: Initial weights and biases of the DNN\\
    \textbf{Output}: Optimized weights and biases of the DNN\\
 
\For{$t\gets1$ \KwTo number of steps}{
    Use weights and biases from previous step\\
     \While{Not minimized}{
        Obtain $\boldsymbol{\hat{u}}$ from the DNN\\
        Compute $\boldsymbol{\nabla} \boldsymbol{\hat{u}}$ using automatic differentiation\\
        Compute $\boldsymbol{\varepsilon}_{t+1}$, $\boldsymbol{e}_{t+1}$, $\boldsymbol{s}_{t+1}^{trial}$, $\boldsymbol{\eta}_{t+1}^{trial}$, and $y_{t+1}^{trial}$\\
        \uIf{$y_{t+1}^{trial}\leq 0$}{
        $\left(\cdot\right)_{t+1}=\left(\cdot\right)^{trial}_{t+1}$\\
        }
        \uElse{
        Compute $\boldsymbol{n}_{t+1}$, $\Delta\boldsymbol{\gamma}_{t+1}$, and $\alpha_{t+1}$\\
        }
        Compute $\boldsymbol{q}_{t+1}$, $\boldsymbol{e}^{p}_{t+1}$, and $\boldsymbol{\sigma}_{t+1}$\\
        \uIf{$\boldsymbol{X}_{int}$}{
        Compute $\boldsymbol{\nabla}\cdot \boldsymbol{\sigma}$ using automatic differentiation\\
        Calculate $MSE_G$\\
        }
        \uElseIf{$\boldsymbol{X}_{t}$}{
        Compute $\boldsymbol{t}= \boldsymbol{\sigma}\cdot\boldsymbol{n}$\\
        Calculate $MSE_t$\\
        }
        \uElse{
        Calculate $MSE_u$\\
        }
        Calculate loss function\\
        Update the weights and biases\\
     }
}     
\caption{von Mises plasticity with isotropic and kinematic hardening}
\label{algo_plast}
\end{algorithm}

\begin{figure}[!htb]
    \centering
    \includegraphics[width=1\textwidth]{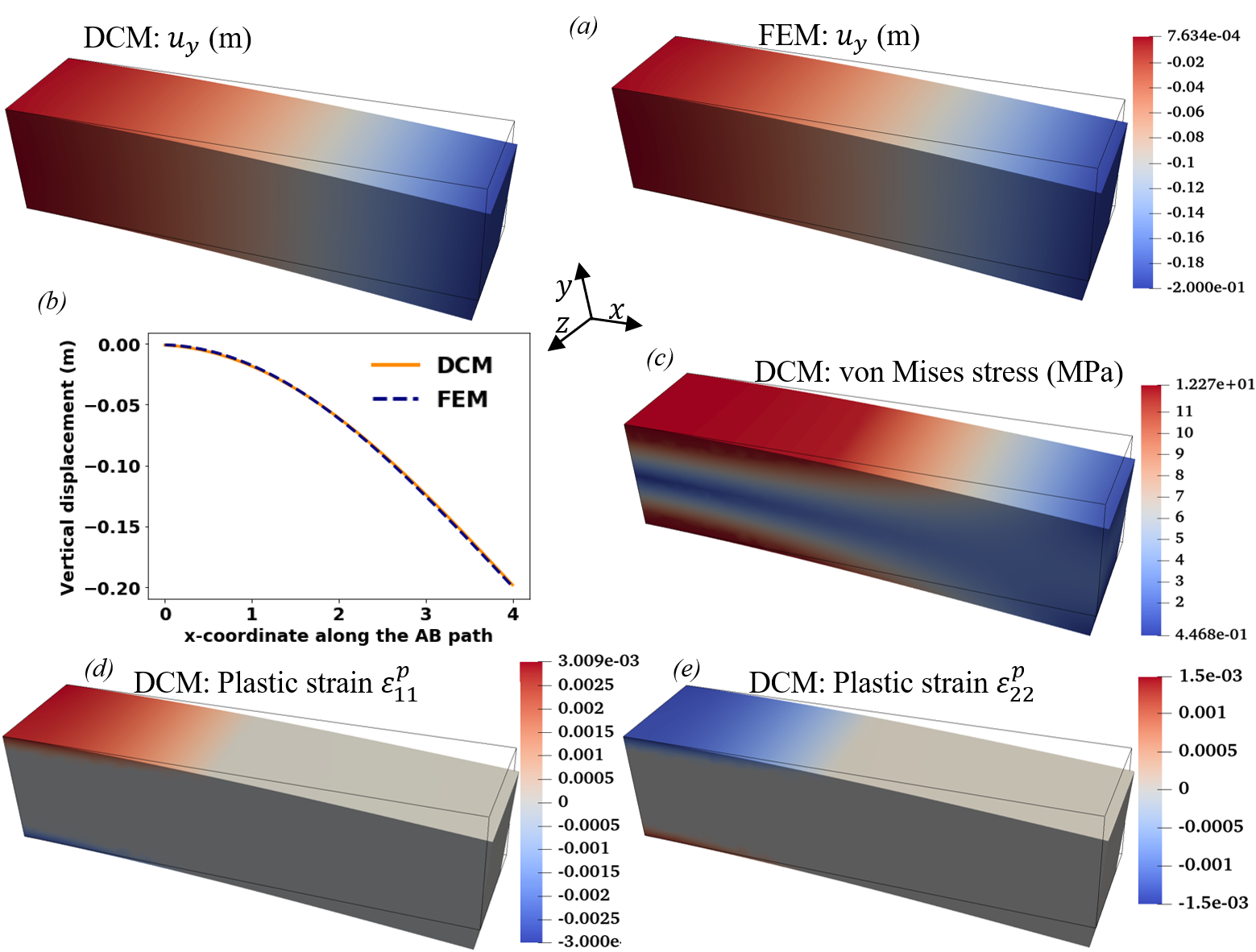}
    \caption{Plasticity example: (a) Comparison between the vertical displacement $\hat{u}_y$ contours obtained from the DCM and FEM, (b) Comparison between the DCM and FEM vertical displacement $\hat{u}_y$ along the $AB$ path, and (c) von Mises stress contours, (c) plastic strain $\varepsilon_{11}^p$, and (d) plastic strain $\varepsilon_{22}^p$. All results are at the last load increment, i.e., $C=0.2\;\text{m}$.}
    \label{PlasticityCB}
\end{figure}

We use the DCM to solve the cantilever beam problem, shown in Figure \ref{Can_BM} with $C=0.2\; \text{m}$, where the beam is made of an elastoplastic material. The numbers of points sampled from the cantilever beam domain are as follows: $N_G=7500$, $N_u=4000$, and $N_t=4000$. Figure \ref{PlasticityCB} portrays the vertical displacement $\hat{u}_y$ contours obtained using the DCM and FEM, and also it shows $\hat{u}_y$ along the $AB$ path. Additionally, we present the von Mises stress and some plastic strain components contours in Figure \ref{PlasticityCB}. Elastoplastic materials are path-dependent. Even if the problem is rate-independent, a pseudo-time stepping scheme is implemented, where the state variables at a step $t$ are inputs for the next step $t+1$. Hence, unlike the linear elasticity and large deformation hyperelasticity, resampling the points at each optimization iteration is tricky. Then, one needs to adopt interpolating techniques to find the state variables at the resampled points, where such interpolation functions might jeopardize accuracy.

\section{Discussion, conclusions, and future directions}\label{conclu}

In this paper, the collocation method and deep learning are merged here to solve partial differential equations involved in the mechanical deformation of different materials: linear elasticity, neo-Hookean hyperelasticity, and von Mises plasticity with isotropic and kinematic hardening. This approach is not data-driven in the sense that no data generation is required. Data generation is usually the most time-consuming stage in developing a data-driven model. However, physics laws are used to obtain the solution. Once the DNN model is appropriately trained, high-quality solutions can be inferenced almost instantly for any point in the domain based on its spatial coordinates. The deep collocation method is meshfree. Thus, there is no need for defining connectivity between the nodes and mesh generation, which can be the bottleneck in many cases \cite{bourantas2018strong} and may require special partitioning methods for large meshes \cite{borrell2018parallel}. Also, meshfree methods have additional advantages, such as there is no element distortion or volumetric locking to worry about.

Using deep learning to solve partial differential equations is a relatively recent research topic, and several issues are still open for improvement. The optimization of a deep learning model is a nonconvex one in most cases. Hence, one needs to be vigilant of getting trapped in local minima. Also, using more intricate methods to identify the optimal activation functions, architecture, and hyperparameters of a DNN model is another active research area; one possible option is to use genetic algorithms \cite{hamdia2020efficient}. In this study, we have used Monte Carlo sampling. It would be interesting to investigate the effect of sampling techniques (such as Latin hypercube, Halton sequence, etc.) and explore whether different sampling techniques affect the accuracy and figure out which ones lead to higher accuracy. Also, resampling the points at each optimization iteration, especially for linear elasticity and hyperelasticity, is another direction to explore. 

In this study, we assumed that the body and inertial forces are negligible. It would be interesting to include such effects in future works and investigate how they impact the model's accuracy. Moreover, for all the cases studied here, we have used feedforward neural networks. However, one can use more advanced architectures such as long short-term memory (LSTM), gated recurrent unit (GRU), or temporal convolutional network (TCN), which are deep sequence learning model. Such architectures are very useful for cases that are history-dependent such as plasticity and viscoplasticity \cite{mozaffar2019deep, abueidda2020deep}. Therefore, it would be intriguing to employ such sequence learning models within the DCM to solve path-dependent problems. Additionally, in future works, other geometries, including irregular geometries, need to be considered and study the DCM's accuracy for more complex geometries. It is also worth mentioning that the penalty method has been used to satisfy the constraints (boundary conditions). One can reformulate the loss function, so the constraints (boundary conditions) are satisfied using Lagrange multipliers. The displacement field and Lagrange multipliers can be inferenced by the DNN, and then one can check if a reasonable accuracy can be obtained. Solving partial differential equations using deep learning methods is an active research area, and this study is far from being the last word on the topic. 
\section*{Acknowledgment}
The authors would like to thank the National Center for Supercomputing Applications (NCSA) Industry Program and the Center for Artificial Intelligence Innovation.
\section*{Data availability}
The data that support the findings of this study are available from the corresponding author upon reasonable request. 

\bibliography{mybibfile}

\end{document}